\documentclass[sigconf]{acmart}
\usepackage{booktabs}
\usepackage{amssymb}
\usepackage{multirow}

\copyrightyear{2021}
\acmYear{2021}
\setcopyright{acmcopyright}\acmConference[MM '21]{Proceedings of the 29th ACM International Conference on Multimedia}{October 20--24, 2021}{Virtual Event, China}
\acmBooktitle{Proceedings of the 29th ACM International Conference on Multimedia (MM '21), October 20--24, 2021, Virtual Event, China}
\acmPrice{15.00}
\acmDOI{10.1145/3474085.3481544}
\acmISBN{978-1-4503-8651-7/21/10}

\begin{document}
\fancyhead{}

\title{Focusing on Persons: Colorizing Old Images Learning from Modern Historical Movies}


\author{Xin Jin, Zhonglan Li, Ke Liu}
\affiliation{%
  \institution{Department of Cyber Security, Beijing Electronic Science and Technology Institute}
  \streetaddress{No 7, Fufeng Road}
  \city{Fengtai District}
  \state{Beijing 100070}
  \country{China}
}

\author{Dongqing Zou}
\authornote{Corresponding author}
\email{zoudongqing@sensetime.com}
\affiliation{%
  \institution{SenseTime Research and Tetras.AI}
  \city{Beijing 100080}
  \country{China}
}
\affiliation{%
  \institution{Qing Yuan Research Institute, Shanghai Jiao Tong University}
  \city{Shanghai 200240}
  \country{China}
}

\author{Xiaodong Li, Xingfan Zhu, Ziyin Zhou, Qilong Sun, Qingyu Liu}
\affiliation{%
  \institution{Department of Cyber Security / Cryptography, Beijing Electronic Science and Technology Institute}
  \streetaddress{No 7, Fufeng Road}
  \city{Fengtai District}
  \state{Beijing 100070}
  \country{China}
}

\renewcommand{\shortauthors}{X. Jin, Z. Li, K. Liu, D. Zou, et al.}


\begin{abstract}
In industry, there exist plenty of scenarios where old gray photos need to be automatically colored, such as video sites and archives. In this paper, we present the HistoryNet focusing on historical person's diverse high fidelity clothing colorization based on fine grained semantic understanding and prior. Colorization of historical persons is realistic and practical, however, existing methods do not perform well in the regards. In this paper, a HistoryNet including three parts, namely, classification, fine grained semantic parsing and colorization, is proposed. Classification sub-module supplies classifying of images according to the eras, nationalities and garment types; Parsing sub-network supplies the semantic for person contours, clothing and background in the image to achieve more accurate colorization of clothes and persons and prevent color overflow. In the training process, we integrate classification and semantic parsing features into the coloring generation network to improve colorization. Through the design of classification and parsing subnetwork, the accuracy of image colorization can be improved and the boundary of each part of image can be more clearly. Moreover, we also propose a novel Modern Historical Movies Dataset (MHMD) containing 1,353,166 images and 42 labels of eras, nationalities, and garment types for automatic colorization from 147 historical movies or TV series made in modern time. Various quantitative and qualitative comparisons demonstrate that our method outperforms the state-of-the-art colorization methods, especially on military uniforms, which has correct colors according to the historical literatures.
\end{abstract}



\begin{teaserfigure}
  \includegraphics[width=\textwidth]{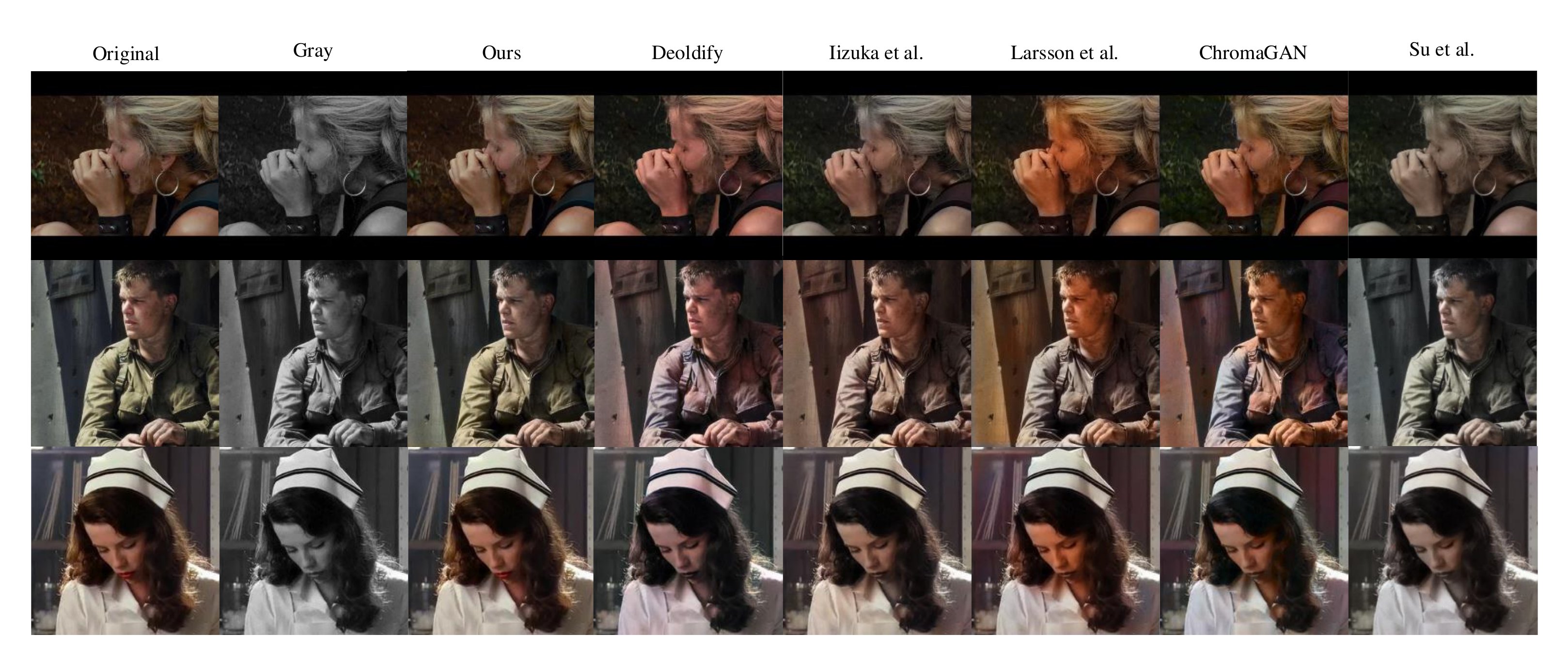}
  \caption{We propose a colorization method focusing on persons that considers fine grained semantic parsing and correct color of images. Our method performs better on historical images.}
  \label{shoutu}
\end{teaserfigure}


\maketitle

\section{Introduction}

Image colorization is a classic image editing problem and has always been a research hotpot in the field of image editing. Many black-and-white military photos were left in the war years. The color restoration of these photos can better
understand the history and have great significance. Also, automatic colorization has a very wide range of applications in industry. There are a large number of images that need to be colored with modern technology to present better visual effects in some video websites such as YouTube. Many archives and museums also have similar requirements. However, so many colorization methods are not suitable for repairing these photos, they do not consider the fine grained semantic parsing of military uniform and the classification of clothing, which makes the coloring result poor. In this paper, the superiority of HistoryNet network to the colorization of person's clothing is shown mainly through the colorization of military uniform, which has more practical significance.

Traditional colorization methods \cite{bahng2018coloring,chen2018language,liu2017recurrent,manjunatha2018learning} propose methods with semantics of input text and language description. Although these methods obtain the semantic segmentation information of reference images, they ignore the boundary of person and background, and also don't consider the human parsing of each part. With the advances of deep learning, many researchers use CNN or GAN to extract information of grayscale images for colorization such as \cite{iizuka2016let,mouzon2019joint,larsson2016learning, isola2017image, nazeri2018image, cao2017unsupervised, vitoria2020chromagan, yoo2019coloring}. These methods often realize the natural color matching, which means the colorization is a subjective problem. However, the colorization of historical military uniform is realistic and practical. In addition, most of the existing modern color image datasets lack the old content or information of real gray historical images, especially the colors of the garments of the historical persons. This will also lead to the poor performance of current colorization methods in military uniforms, as shown in the second line of Figure \ref{shoutu}.

To address these shortcomings, we propose a new HistoryNet architecture and a MHMD dataset. First, there are a great variety of military uniform in history and we can obtain the eras, nationalities and garment types information of an image. According to these, we design the labels of images and classification subnetwork. Classification submodule supplies classifying of images according to the eras, nationalities and garment types labels. In addition, we have designed a classifier submodule in discriminator, which joint classification subnetwork to constrain loss for getting more suitable colorization. Inspired by InfoGAN \cite{2016InfoGAN}, we have designed classifier subnetwork that can better present the classification information. Second, we have designed fine grained semantic parsing subnetwork, which supply the semantic understanding of historical person's clothing. Through this subnetwork, the historical person's each part (such as face, arms, hair and so on) will be separate and obtain a clear boundary finally. In the training process, based on U-net \cite{ronneberger2015u}, we connect the up-sample layers of image features on the parsing subnetwork with the generator network of colorization. And in the future, we also hope to apply relevant research methods to video colorization. Third, we propose a new dataset called Modern Historical Movies Dataset (MHMD) for automatic image colorization. We randomly select 2,000 images from ImageNet \cite{deng2009imagenet} and MHMD datasets to caculate their average values on H channel of HSV.  MHMD dataset focus on blue and red yellow of color area and ImageNet dataset is distributed in all color areas, especially in yellow and blue areas. This shows that MHMD dataset considers the old content and information of real gray historical images, which are mostly dark blue and red yellow. However, the distribution of ImageNet dataset is more balanced in H value. Therefore, our method can perform well in colorizing old images. Finally, the contributions of this work include:



\begin{enumerate}
	\item Semantic parsing subnetwork can accurately obtain the semantic information of various parts of person, which makes the image colorization boundary and the human semantic parsing more accurate.
	
	\item In HistoryNet, we have designed classifier and classification subnetworks. Info information of classifier subnetwork and classification labels jointly achieve HistoryNet coloring accuracy and classification labels more precise.
	
	\item We propose a dataset of persons colorization which is called MHMD: Modern Historical Movies Dataset. MHMD contains 1,353,166 images and the 42 labels of eras, nationalities, and garment types, which have great influences on the colorization of old images.
\end{enumerate}

\section{Related work}
\textbf{Reference Image-based Methods} Manga also called Japanese comic is popular all over the world. Most of them are monochrome. Therefore, many manga colorization methods appeared which are based on a reference image of sketch and line art. \cite{sato2014reference} propagate the colors of the reference manga to the target manga by representing images as graphs and matching graphs. \cite{zhang2017style} integrate residual U-net to apply the style to the grayscale sketch with auxiliary classifier generative adversarial network (AC-GAN) \cite{pmlr-v70-odena17a}.  \cite{ci2018user} propose a novel deep conditional adversarial architecture for scribble based anime line art colorization. \cite{hensman2017cgan} propose a manga colorization method based on conditional Generative Adversarial Networks (cGAN) and require only a single colorized reference image for training. \cite{furusawa2017comicolorization} colorize a whole page (not a single panel) semiautomatically, with the same color for the same character across multiple panels. In addition, \cite{welsh2002transferring,xian2018texturegan} focus on the texture and luminance information of the reference image to achieve colorization. \cite{ironi2005colorization,sun2019adversarial,zhang2017real} are other methods based on reference image. \cite{ironi2005colorization} colorize images from pixel level. \cite{sun2019adversarial} propose a dual conditional generative adversarial network which considers contour and color style of images.

\textbf{Semantics-based Methods} Researchers have propose many semantic-based methods to deal with the colorization problem of grayscale images. \cite{bahng2018coloring,chen2018language,liu2017recurrent,manjunatha2018learning} colorize gray images based on semantics of input text and language description. \cite{zou2019language,zou2018sketchyscene} both propose a method based on scene sketches and semantic segmentation. \cite{zhao2019pixelated,su2020instance} learn object-level semantics to guide image colorization. \cite{zhao2019pixelated} propose to exploit pixelated object semantic to guide image colorization, which also consider the semantic categories of objects. \cite{deshpande2015learning} achieve auto image colorization by learning from examples. \cite{kim2019tag2pix} propose a Tag2Pix GAN architecture which takes a grayscale line art and color tag information as input to produce a quality colored image.

\textbf{Deep Learning-based Methods} Cheng et al. \cite{cheng2015deep} first propose a neural network method for automatic colorization. With the development of CNN, \cite{iizuka2016let,mouzon2019joint,larsson2016learning} use CNN to extract information of images. In the work of Iizuka et al. \cite{iizuka2016let}, the colorization network can obtain both local and global features of the image. In the research of Mouzon et al. \cite{mouzon2019joint}, the distribution statistical method is combined with the variational method to calculate the possible color probability for each pixel of the image. \cite{larsson2016learning} train a VGG to learn the color histogram of each pixel.

Many existing GAN have the ability to learn the probability distribution of high dimensional spatial data, which can be applied to colorization tasks. The method in \cite{isola2017image} use conditional GAN to map the input grayscale image to the output colorized image. Nazeri et al. \cite{nazeri2018image} attempt to fully generalize the colorization procedure using a conditional DCGAN. \cite{cao2017unsupervised} leverage the conditional GAN to automatically obtain a variety of possible colorization results through multiple sampling of the input noise. \cite{vitoria2020chromagan} propose an adversarial learning colorization approach coupled with semantic information. \cite{yoo2019coloring} present a novel memory-augmented colorization model Memo-Painter that can produce high-quality colorization with limited data.


\section{Architecture and Training Losses}

\begin{figure*}[htbp]
	\centering
	\includegraphics[width=\linewidth]{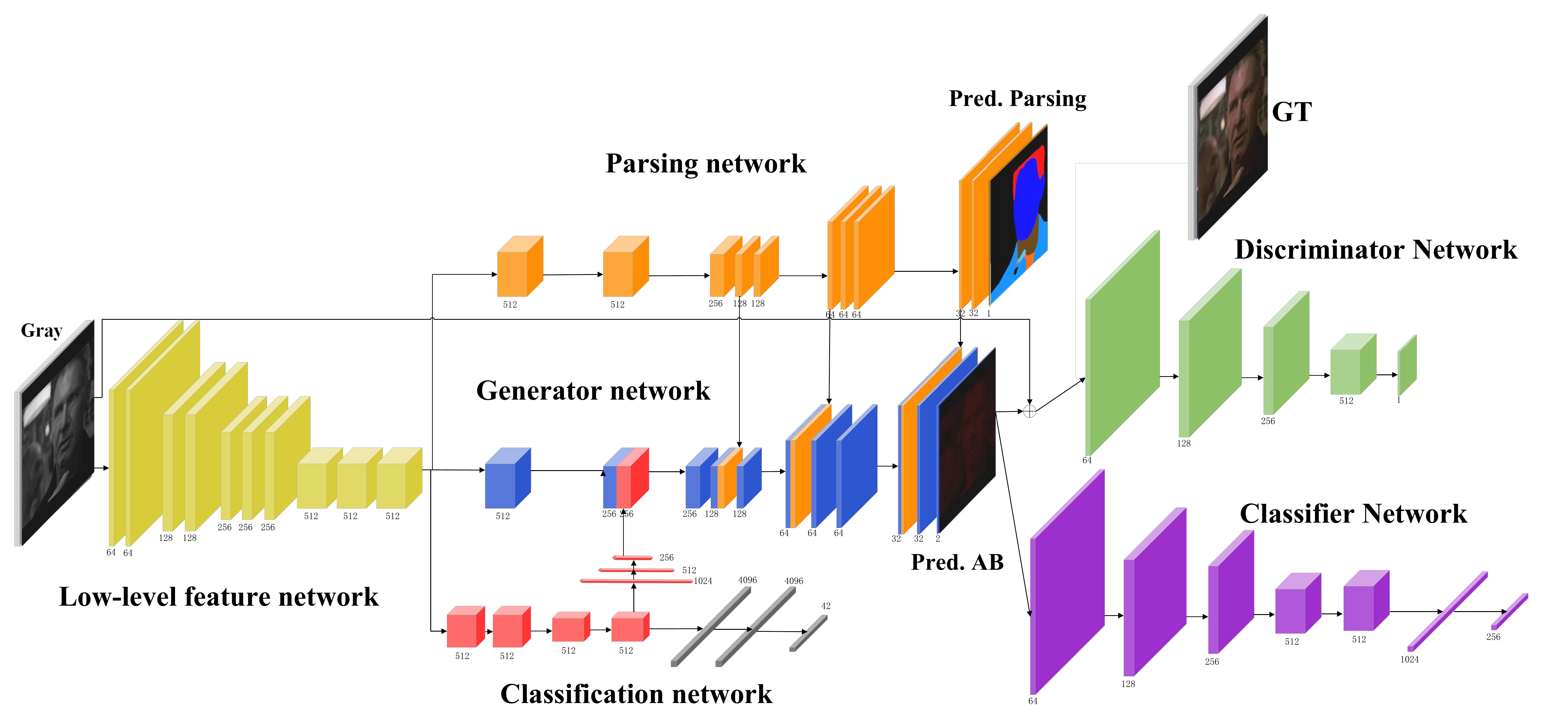}
	\caption{HistoryNet network structure diagram. Overview of our model, it combines a low-level feature network $G_{0}$ (in yellow), a Generator network $G_{1}$ (include yellow, blue, red and orange layers), a classification network $G_{2}$ (in red), a parsing network $G_{3}$ (in orange) and a Discriminator network. The discriminator network is divided into two parts which are $D_{1}$ (in green) and $D_{2}$ (in purple).}
	\label{network}
\end{figure*}

According to the description of the HistoryNet network structure, we define the total loss function as:
\begin{equation}
	L\left(G_{\theta },D_{w}\right)=L_{r}\left(G_{\theta _{1}}^{1}\right)+\lambda _{cls}L_{cls}\left(G_{\theta _{2}}^{2}\right)
\end{equation}
\begin{equation*}
	+\lambda _{par}L_{par}\left(G_{\theta _{3}}^{3}\right)+{\lambda _{g}}{L_{g}}\left(G_{\theta _{1}}^{1},D_{w}\right)+\lambda _{info}L_{info}\left(G_{\theta _{1}}^{1}\right)
\end{equation*}

The first three terms in the formula are the loss values of the generator, we denote them by $G_{\theta}=(G_{\theta _{1}}^{1}, G_{\theta _{2}}^{2}, G_{\theta _{3}}^{3})$, where $\theta=(\theta _{1}, \theta _{2}, \theta _{3})$ stand for all the generator parameters. The last two terms are the loss values of the discriminator which we denote by $ D_{w}$.

\textbf{Generator Network} The generator network mainly generates $(a, b)$ channel images. We define its loss function as:

\begin{equation}
	L_{r}\left(G_{\theta _{1}}^{1}\right)=E_{\left(L,{a_{r}},{b_{r}}\right)\sim {P_{r}}}\left[||G_{\theta _{1}}^{1}\left(L\right)-\left(a_{r},b_{r}\right)||_{2}^{2}\right]
\end{equation}

Where $(L,a_{r},b_{r})$ is the representation of the color image in the CIE $L_{ab}$ color space, and $P_{r}$ is the distribution of the color image. $||\cdot ||_{2}$ is the Euclidean distance. By calculating the Euclidean distance between $G_{\theta _{1}}^{1}\left(L\right)$ and $\left(a_{r},b_{r}\right)$, the resulting image can better perceive the color difference with the real image in the $L_{ab}$ color space. $L2$ loss has enough ability to constrain the network to obtain more realistic colorization results.

\textbf{Parsing Network} While many methods consider the semantic segmentation for colorization, they are not suitable for the historical person's diverse clothing colorization, thus we propose fine grained semantic understanding for this issue. Under the guidance of human parsing as ground truth, the parsing feature of the image is obtained by continuously up-sampling and based on U-net \cite{ronneberger2015u}, we concatenate the up-sampling information of parsing network with the generator network (the blue $G_{1}$). Through this way, the generator network can obtain the fine grained semantic parsing information of images. For example, the person's face, hat, hands and clothes are separated to ensure the colorization boundary is clear and accurate. As shown in Figure \ref{Parsing}. The parsing network loss we defined is:

\begin{equation}
	L_{par}\left(G_{\theta _{3}}^{3}\right)=E_{\left(S,{a_{r}},{b_{r}}\right)\sim {P_{r}}}\left[||G_{\theta _{3}}^{3}\left(S\right)-\left(a_{r},b_{r}\right)||_{2}^{2}\right]
\end{equation}

In $(S,a_{r},b_{r})$, $S$ is the image before parsing, and $(a_{r},b_{r})$ is the parsing image. Calculating the Euclidean distance between the image generated by the parsing network and the actual parsing image and minimize it. In this way, the image input to the parsing network is closer to the real image and finally the colorization result is more accurate.

\textbf{Classification network} As shown in red in Figure \ref{network}, classification network is designed to obtain the high-level features of the image and the category label information of the image colorization. The convolution of four modules are then divided into two parts of full connection layers. The gray fully connected layers on the right side obtain a 42-dimensional vector that represents the 42 classification labels in the dataset we developed. 42 labels consider the eras, nationalities and garment types of historical person's images. The detailed labels of classification improve the accuracy of image colorization such as the navy clothing is mainly white. The second part (as shown in the red part) is concatenated together with the blue $G_{1}$ convolution layer above. This part can obtain 256-dimensional vector which is corresponding to classifier network to constrain the loss for correct color classification of images. The loss function is defined as:

\begin{equation}
	L_{cls}\left(G_{\theta _{2}}^{2}\right)=E_{L\sim {P_{rg}}}\left[KL\left(y_{v}||G_{\theta _{2}}^{2}\left(L\right)\right)\right]
\end{equation}

Where $P_{rg}$ is the distribution of the input grayscale image. $y_{v}\in R^{m}$ is the classification vector obtained by the VGG network classifying the images in the dataset, and $m$ is the number of image classifications. $KL$ divergence calculate the loss due to $y_{v}$ fitting $G_{\theta _{2}}^{2}(L)$. The input grayscale image is processed by the classification network so that the coloring network can choose color more correctly.

\textbf{Classifier Network} In general colorization methods, the loss of colorization is required to be minimized as much as possible, so that the colorization result is closer to the real image. For example, when we want to colorize a certain object blue sky, no matter how many images are trained, the final average color value is blue so that the colorization image is still blue of sky. However, due to the nationality variations, garment types and era differences, these factors will hinder the calculation of color loss value of historical person's images especially military uniform. According to the traditional method of calculating the minimum loss between the colorization image and the real image, the color of the final colorization image will tend to be gray after averaging. So, classifier network adopt InfoGAN \cite{2016InfoGAN}, which can make good use of the information of image latent code to achieve better performance of images. After obtaining the $(a, b)$ channel of image through the generator, we input it into the classifier network, and finally obtain a 256-dimensional vector through the fully connected layers. The $KL$ divergence between the 256-dimensional vector obtained in classification network $G_{2}$ is expected to minimize the fitting loss so that the generated image can better present the classification information. 

During this way, final colorization images can have correct color for each part of garment. In addition, the interior coloring of each part of the garment is continuous. As shown in Figure \ref{ablation}. The loss we designed is:

\begin{equation}
	L_{info}\left(G_{\theta _{1}}^{1}\right)=E_{L\sim {P_{rg}}}\left[KL\left(\textit{infoGAN}||G_{\theta _{1}}^{1}\left(L\right)\right)\right]
\end{equation}

\textbf{Discriminator Network} As shown in Figure \ref{network}, $D_{1}$ is based on the Markov discriminator architecture. The PatchGAN \cite{isola2017image} discriminator can track and capture the high-frequency structure of the generated image, thereby making up for the high-frequency information loss caused by the use of $L2$ loss in $G_{1}$. For this reason, we define each patch as true or fake.

In this paper, based on WGAN \cite{arjovsky2017wasserstein}, we design the loss of discriminator. WGAN \cite{arjovsky2017wasserstein} uses the Earth-Mover distance to minimize the possible and true distribution of the generator. Using this feature of the WGAN \cite{arjovsky2017wasserstein} network can avoid gradient disappearance and mode collapse during the training process and eventually achieve stable training and obtain better-colored images. Also, we use Kantorovich - Rubinstein duality \cite{kantorovitch1958translocation,villani2008optimal} and add the gradient penalty term to constrain the $L2$ norm of the discriminator relative to its input, thus defining $D_{w}\in D$, where $D$ denotes the set of 1-Lipschitz functions.

\begin{equation}
	\begin{split}
		L_{g}\left(G_{\theta _{1}}^{1},D_{w}\right)&=
		E_{\tilde{I}\sim {P_{r}}}\left[D_{w}\left(\tilde{I}\right)\right]\\
		&-E_{\left(a,b\right)\sim {P_{G_{\theta_{1}}^{1}}}}\left[D_{w}\left(L,a,b\right)\right]\\
		&-E_{\hat{I}\sim {P_{\hat{I}}}}\left[\left({{||\nabla_{\hat{I}}}{D_{w}}\left(\hat{I}\right)||_{2}}-1\right)^{2}\right]
	\end{split}
\end{equation}

Where $P_{G_{\theta _{1}}^{1}}$ is the model distribution in $G_{\theta _{1}}^{1}(L)$ of L $\sim$ $P_{rg}$. As in \cite{gulrajani2017improved}, $P_{\hat{I}}$ is a straight-line uniform sampling which is between the pairs of points sampled along the data distribution $P_{r}$ and the generator distribution $P_{G_{\theta _{1}}^{1}}$.  

According to the formula (1) (total loss function) proposed above, we train the networks $G$ and $D$ by calculating the following expressions. During the training process, We take the hyperparameter values $\lambda _{cls},\lambda _{par},\lambda _{g},\lambda _{info}$ as: 0.003, 0.003, 0.1, 0.003.

\begin{equation}
	\min _{G_{\theta }} \max _{D_{w}\in D} L\left(G_{\theta },D_{w}\right)
\end{equation}

\section{Datasets}

\begin{figure}[htbp]
	\centering
	\includegraphics[width=\linewidth]{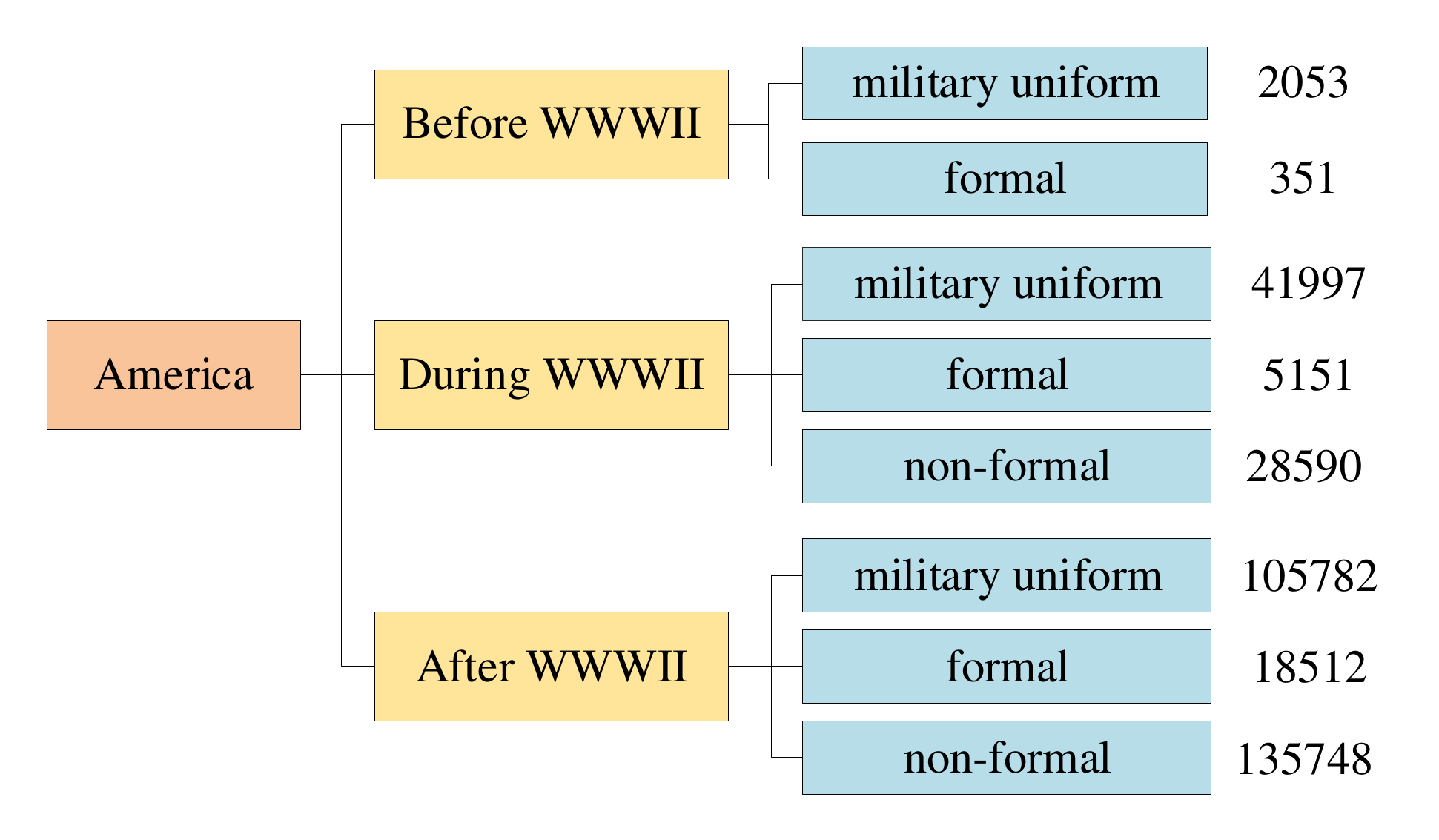}
	\caption{The picture shows the U.S. military uniform, formal and non formal labels and the corresponding number of pictures before, during and after World War II in MHMD.}
	\label{yuyishu}
\end{figure}

The colorization of historical person's clothing is of practical significance, especially the colorization of historical military uniform. At present, most of the existing modern color image datasets contain modern objects or scenes but lack old content or information of real gray historical images, especially the colors of the garments of the historical persons. Therefore, we build a dataset called MHMD: Modern Historical Movies Dataset. MHMD can meet the requirements of garment types, eras and nationalities. First of all, we search for 147 historical movies and TV series in modern time. Second, after preprocessing, the MHMD dataset obtains 1,353,166 images,including 1.2M images on training dataset and 100001 images on testing dataset. We classify the images into 42 labels according to above types, as shown in Figure \ref{yuyishu}. The 42 labels are divided by us have solved the problem of different military uniforms of different countries in different periods. Detailed classification of labels is helpful to obtain accurate information for network structure. On this basis, we design the HistoryNet network structure to realize the colorization of historical military uniforms and the restoration of old historical photos. More details about MHMD can be found on https://github.com/BestiVictory/MHMD. 

\subsection{Collecting Methods}

We present MHMD of approximately 1.3M images with image-level labels and localization instructions. We mainly download TV series, movies and documentaries from various websites. Because we are focusing on the colorization of historical person's images and in order to ensure the quality of the films (such as definition, etc.), we have formulated the film selection criteria: color films taken after the 1990s, the content of the film is mainly about history, war, person and objects of various countries before 1990. 

\subsection{Processing Methods}

First, we cut the film into images, and use the program to remove black and white images, images with too low pixel blur. For the goal of focusing on persons, we use Yolov3 to judge whether there is a person in the image and delete the images without people or too many people. Through these preprocessing, the diversity of the content and the clarity of the image are guaranteed. Secondly, through an image, we can obtain the eras, nationalities and garment types information of it. According to this, we have designed 42 kinds of labels and divide them into three categories: era, nationality and garment type. The era label is divided into: before, during and after World War II. Nationality labels are: China (divided into the Communist Party and the Kuomintang), Japan, the United States, Germany, Britain and Russia. There are three types of garment type: military, formal (such as suits) and informal. Thirdly, We randomly extract 1\% of the images from the dataset, which is the same distribution as the original dataset, and then we label the 1\% images manually. The 1\% images and data augmentation of these images (such as horizontal image mirroring and Gaussian blur) will be inputted to ResNet \cite{he2016deep} for training. After the training, we input other images in the dataset into ResNet \cite{he2016deep} for automatic classification to obtain labels. The accuracy of classification can reach 98\% or more. The categories of MHMD and the comparison with other colorization datasets are shown in table \ref{dataset}.

\begin{table*}[htbp]
	\caption{Comparison of MHMD with other colorization datasets}
	\label{dataset}

	\begin{tabular}{c|c|c|c|c|c|c|c|c}
		\hline
		& \multicolumn{7}{c|}{The labels of the dataset}                                                                                                                                &                            \\ \hline
		Dataset               & Scene              & \multicolumn{2}{c|}{Era}                                       & \multicolumn{2}{c|}{Nationality} & \multicolumn{2}{c|}{Garment Type}                    & Total \\ \hline
		\multirow{5}{*}{MHMD} & \multirow{5}{*}{×} & \multirow{2}{*}{Before WWII} & \multirow{2}{*}{66,900} & Chinese   & 753,473 & \multirow{2}{*}{Military} & \multirow{2}{*}{707,771} & \multirow{5}{*}{1,353,166}   \\ \cline{5-6}
		&                    &                                      &                         & American         & 934,415        &                           &                          &                            \\ \cline{3-8}
		&                    & \multirow{2}{*}{During WWII} & \multirow{2}{*}{547,318} & Russian          & 45,291         & \multirow{2}{*}{Formal}   & \multirow{2}{*}{104,763}   &                            \\ \cline{5-6}
		&                    &                                      &                         & German           & 59,015         &                           &                          &                            \\ \cline{3-8}
		&                    & \multirow{2}{*}{After WWII}  & \multirow{2}{*}{738,948} & Japanese            & 110,562         & \multirow{2}{*}{Informal} & \multirow{2}{*}{540,632}   &                            \\ \cline{5-6}
		&                    &                                      &                         & English  & 46,641        &                           &                    &                            \\ \hline
		ImageNet \cite{deng2009imagenet}              &  \checkmark                  & \multicolumn{2}{c|}{×}                                         & \multicolumn{2}{c|}{×}           & \multicolumn{2}{c|}{×}                               & about 1,300,000              \\ \hline
		COCO-Stuff \cite{caesar2018coco}           & \checkmark                  & \multicolumn{2}{c|}{×}                                         & \multicolumn{2}{c|}{×}           & \multicolumn{2}{c|}{×}                               & about 164,000               \\ \hline
		Places205 \cite{zhou2014learning}        & \checkmark                  & \multicolumn{2}{c|}{×}                                         & \multicolumn{2}{c|}{×}           & \multicolumn{2}{c|}{×}                               & 20,500                      \\ \hline
	\end{tabular}

\end{table*}

\section{Experiments}

\begin{figure*}[htbp]
	\centering
	\includegraphics[width=\linewidth]{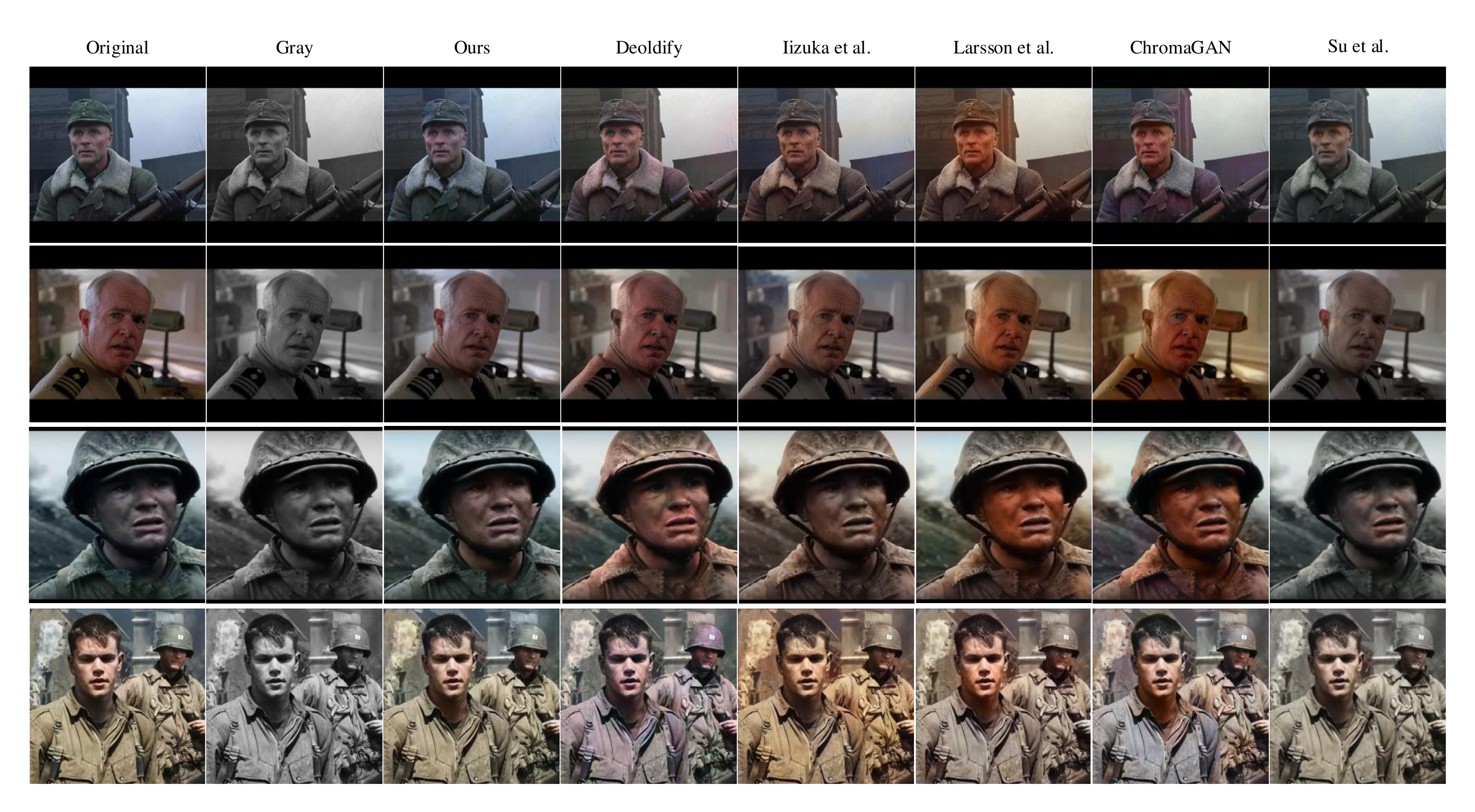}
	\caption{Contrastive experimental diagram. Some qualitative results, from left to right: Original, Gray, Ours, Deoldify \cite{biohit}, Iizuka et al. \cite{iizuka2016let}, Larsson et al. \cite{larsson2016learning}, ChromaGAN \cite{Vitoria_2020_WACV}, Su et al. \cite{su2020instance}. The results are comparable.}
	\label{duibishiyan}
\end{figure*}

In this section, we evaluate our methods quantitatively and qualitatively on MHMD we designed. We compare our experimental results with, Deoldify \cite{biohit}, Iizuka et al. \cite{iizuka2016let}, Larsson et al. \cite{larsson2016learning}, ChromaGAN \cite{Vitoria_2020_WACV}, Su et al. \cite{su2020instance}, and use LPIPS \cite{zhang2018unreasonable}, PSNR and SSIM indicators to quantitatively compare our experimental results with those of the most advanced methods. Finally, we carry out ablation experiments on parsing network and classifier network, and prove that they have positive effects on the improvement of network performance.

\begin{figure*}[htbp]
	\centering
	\includegraphics[width=\linewidth]{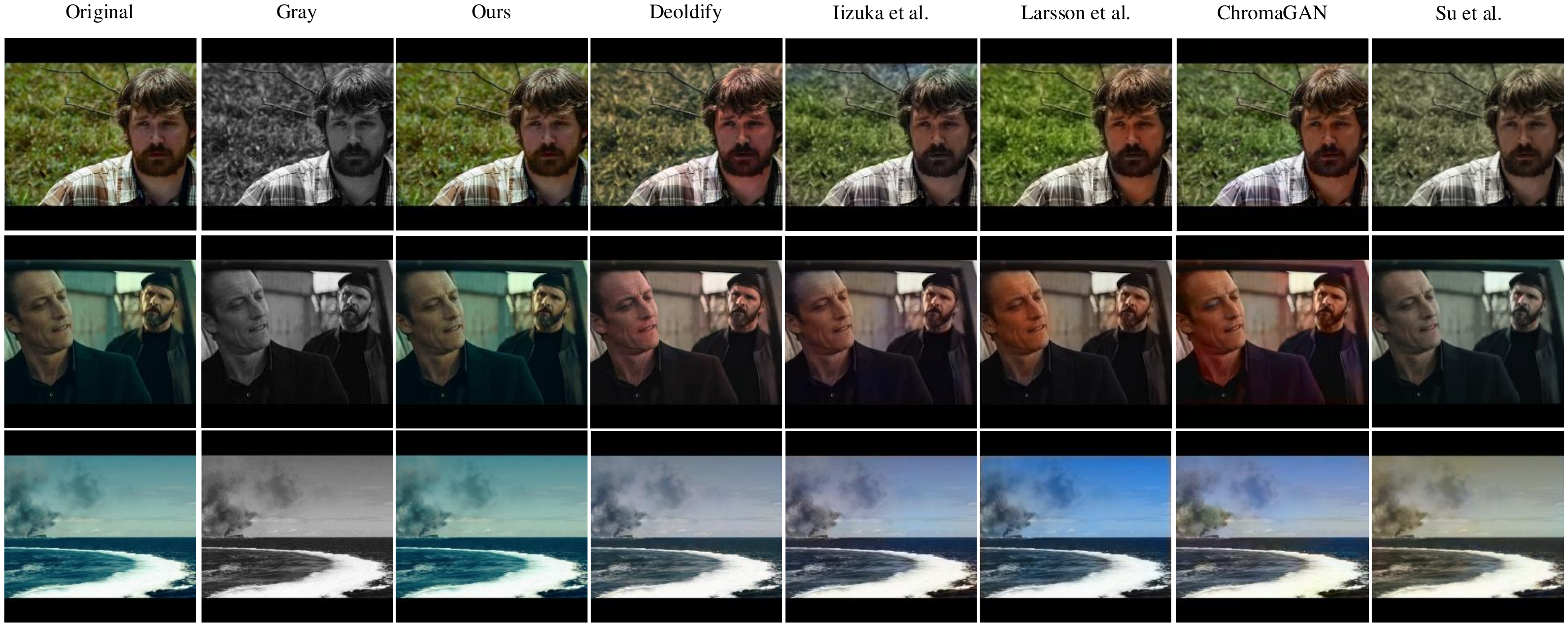}
	\caption{The colorization comparison of natural scenery and person on HistoryNet. Our method also works well in natural scenes.}
	\label{othercomparison}
\end{figure*}

\subsection{Implementation Details}
We train HistoryNet for a total of eight epochs and set the batch size to 16, on the 1.3M images from MHMD resized to 224${\times}$224. A single epoch takes approximately 28 hours on a Nvidia titan X pascal GPU. We minimize our objective loss using Adam optimizer with learning rate equal to 2${\times}$10$^{{-}5}$ and momentum parameters $\beta _{1}=0.5$ and $\beta _{2}=0.999$. We alternate the optimization of the generator $G_{0}$ and discriminator $D_{1}$. The first stage of the network (displayed in yellow in Figure \ref{network}), takes as input a grayscale image of size 224 ${\times}$ 224, and is initialized using the pre-trained weights of the VGG-16 \cite{simonyan2014very} trained on ImageNet.

\subsection{Quantitative Comparisons}
We use three metrics, LPIPS \cite{zhang2018unreasonable}, PSNR and SSIM. LPIPS \cite{zhang2018unreasonable} is a measurement method of perceptual similarity proposed by Zhang et al. that can achieve good perceptual judgment in challenging models of visual prediction and other tasks. The lower the LPIPS \cite{zhang2018unreasonable} value, the smaller the perceived difference between the image and the original image. PSNR and SSIM are two commonly used indexes in image quality evaluation. PSNR is peak signal to noise ratio, which is usually used as an evaluation index of the quality of an image before and after compression. The higher the PSNR value is, the better the quality of the generated image will be. SSIM is structural similarity index, which is an index to measure the similarity between two images. Table \ref{Quantitative} shows the comparison of our experimental results with Deoldify \cite{biohit}, Iizuka et al. \cite{iizuka2016let}, Larsson et al. \cite{larsson2016learning}, ChromaGAN \cite{Vitoria_2020_WACV}, Su et al. \cite{su2020instance}. In the above methods,  We have successfully applied the network structures of ChromaGAN \cite{Vitoria_2020_WACV} and Su et al. \cite{su2020instance} to MHMD for training, and the training details are shown in the supplementary.

\begin{table}[htbp]
	\caption{Quantitative comparison of experimental}
	\label{Quantitative}
	\begin{tabular}{lccc}
		\toprule
		Method & LPIPS↓ & PSNR↑ & SSIM↑\\
		\midrule
		Iizuka et al. \cite{iizuka2016let} & 0.134 & 25.779 & 0.956\\
		Larsson et al. \cite{larsson2016learning} & 0.147 & 24.527 & 0.946\\
		Deoldify \cite{biohit} & 0.127 & 26.321 & 0.957\\
		ChromaGAN \cite{vitoria2020chromagan} & 0.118 & 29.487 & 0.951\\
		Su et al. \cite{su2020instance} & 0.132 & 25.951 & 0.941\\
		\midrule
		HistoryNet & \textbf{0.101} & \textbf{30.638} & \textbf{0.962}\\
		\bottomrule
	\end{tabular}
\end{table}
As can be seen from table 2, our method have better perform in these metrics. The lower LPIPS value indicates that our results are more similar to the source image. The higher PSNR and SSIM values means that the quality of the generated image is better and is similar to the original image.

\subsection{Ablation Experiments}

We have trained ChromaGAN on MHMD dataset as a baseline. As can be seen from table 3, LPIPS, PSNR and SSIM perform better after the parsing and classifier networks are added. This is enough to show that the parsing network plays a positive role in image colorization and classifier network can be better used in image colorization network after obtaining the information of image latent layer. It can be seen from the last row of table 3 that our proposed method can perform well on all three parameters, which fully shows that our method has a positive effect on colorization. We also perform three ablation tests: Baseline + Parsing, Baseline + Classifier and our method.

\begin{table}[htbp]
	\caption{Ablation experiments}
	\label{Ablation}
	\begin{tabular}{lccc}
		\toprule
		Method & LPIPS↓ & PSNR↑ & SSIM↑\\
		\midrule
		ChromaGAN & 0.123 & 27.093 & 0.946\\
		Baseline+Parsing & 0.121 & 28.992 & 0.948\\
		Baseline+Classifier & 0.119 & 29.828 & 0.951\\
		HistoryNet & \textbf{0.107} & \textbf{30.585} & \textbf{0.959}\\
		\bottomrule
	\end{tabular}
\end{table}

\subsection{Qualitative comparisons}

\begin{figure}[htbp]
	\centering
	\includegraphics[width=\linewidth]{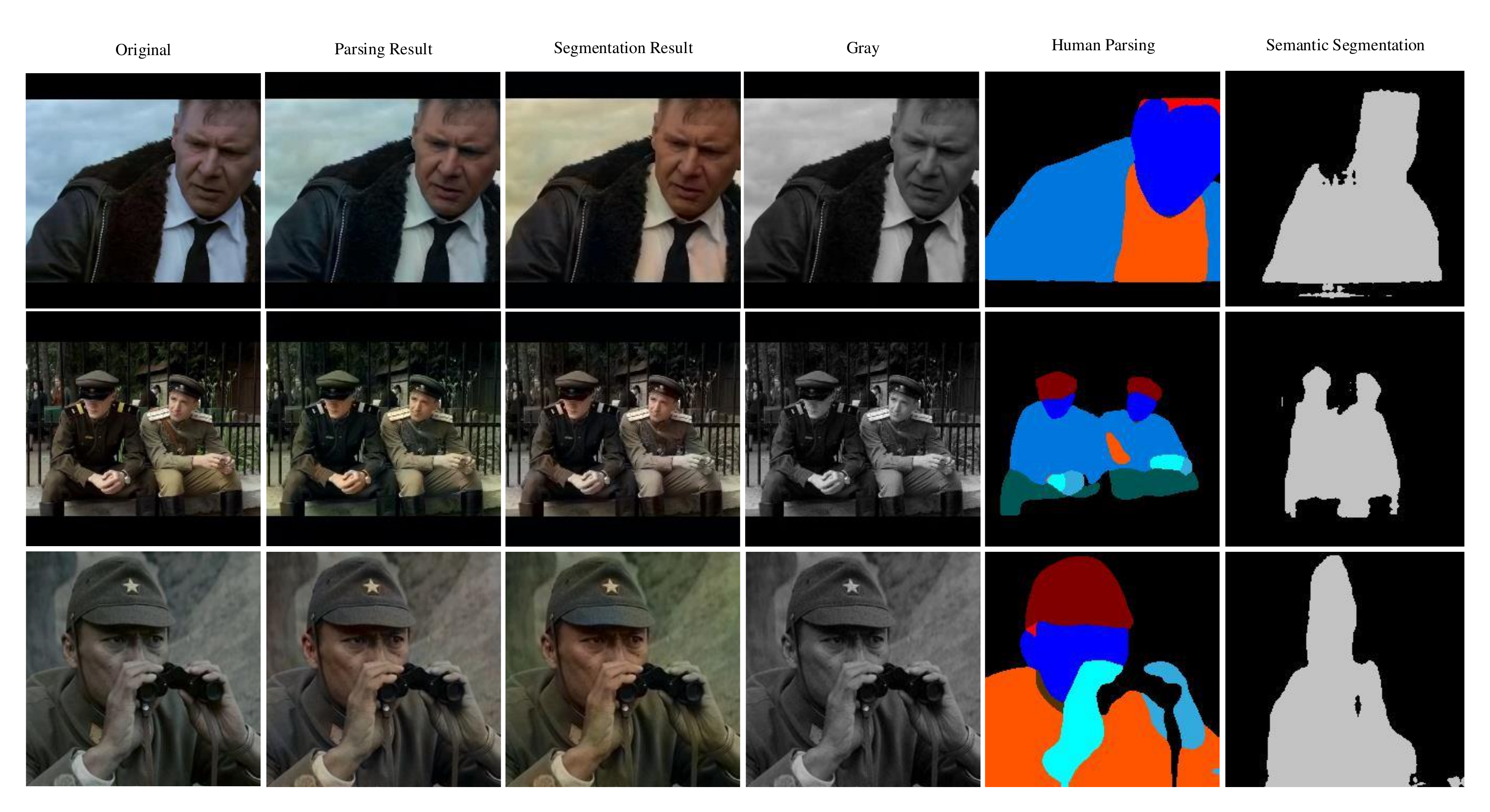}
	\caption{The picture shows that the segmentation of human parsing is more accurate than semantic segmentation.}
	\label{Parsing}
\end{figure}

\textbf{Effect of fine grained semantic parsing} There exist kinds of image segmentation methods in the literatures \cite{gong2018instance,chen2017deeplab,ZhouASOC2020,ZhouSCIS2019,LiTIP2014}. In this paper, we use Human Parsing \cite{gong2018instance} and Deeplab-v3 \cite{chen2017deeplab} separately for semantic segmentation. Deeplab-v3 \cite{chen2017deeplab} only separate persons and background, but can't focus on details of persons' each part, such as face and hands. Therefore, we adopt instance-level human parsing \cite{gong2018instance} for semantic segmentation, which is focus on recognizing each semantic part for example arms and hair. Due to the limit of equipment and time, we separately adjust 75,837 images of human parsing and Deeplab-v3 \cite{chen2017deeplab} manually and use them as the ground truth for parsing network. Through training on HistoryNet, it can be seen in Figure \ref{Parsing} that human parsing focus on fine grained semantic parsing so its segmentation accuracy is better than semantic segmentation. Because of this, the result of using human parsing as ground truth training is better than that of Deeplab-v3. For example, in the third line of the Figure \ref{Parsing}, the person's hands are separated by Human Parsing that the right hand of the person is light green, the left one is light blue. By taking human parsing as the ground truth of parsing network, parsing network can accurately guide the acquisition and colorization of color information of each part. Therefore, the boundary of Parsing Result is more clear and the colorization effect is closer to Original images. However, Semantic Result can not achieve accurate segmentation, which leads to the color fusion of hands and background and presents the same green of the background.

\begin{figure}[htbp]
	\centering
	\includegraphics[width=\linewidth]{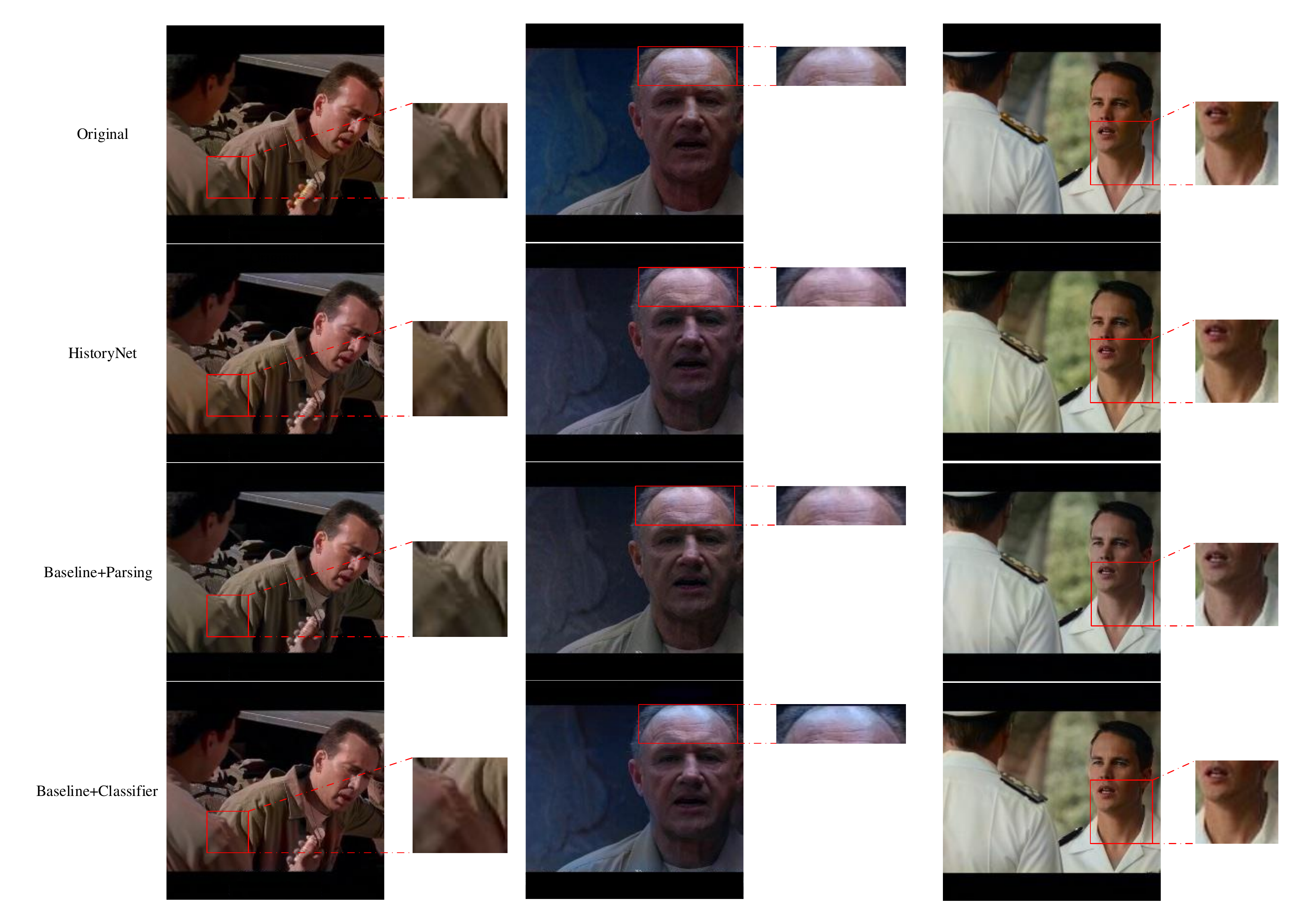}
	\caption{Some qualitative comparisons of ablation experiments. Parsing submodule can obtain the clear boundary through fine grained semantic parsing; Classifier submodule can help choose correct color for colorization.}
	\label{ablation}
\end{figure}

\textbf{Ablation experiments} In HistoryNet, there are two main submodules: parsing and classifier subnetworks, so we do the corresponding ablation experiments. As shown in Figure \ref{ablation}, baseline consists of $G_{0}$, $G_{1}$, $G_{2}$ and $D_{1}$ in Figure \ref{network} of HistoryNet structure. From the Figure \ref{ablation}, we can see that parsing can better segment the boundary, so as to make the colorization more accurate. For example, in the first column, Baseline+Parsing can accurately separate the two clothes and the boundary of the clothes is more precise, while Baseline+Classifier does not perform due to the lack of fine grained semantic parsing of person. Although the parsing results have better boundary of person, the color of the images is not correct and nature, such as the third column in the figure. Compared with Baseline+Classifier, Baseline+Parsing is not classified, so the color of the neck is not natural.

\begin{figure}[htbp]
	\centering
	\includegraphics[width=\linewidth]{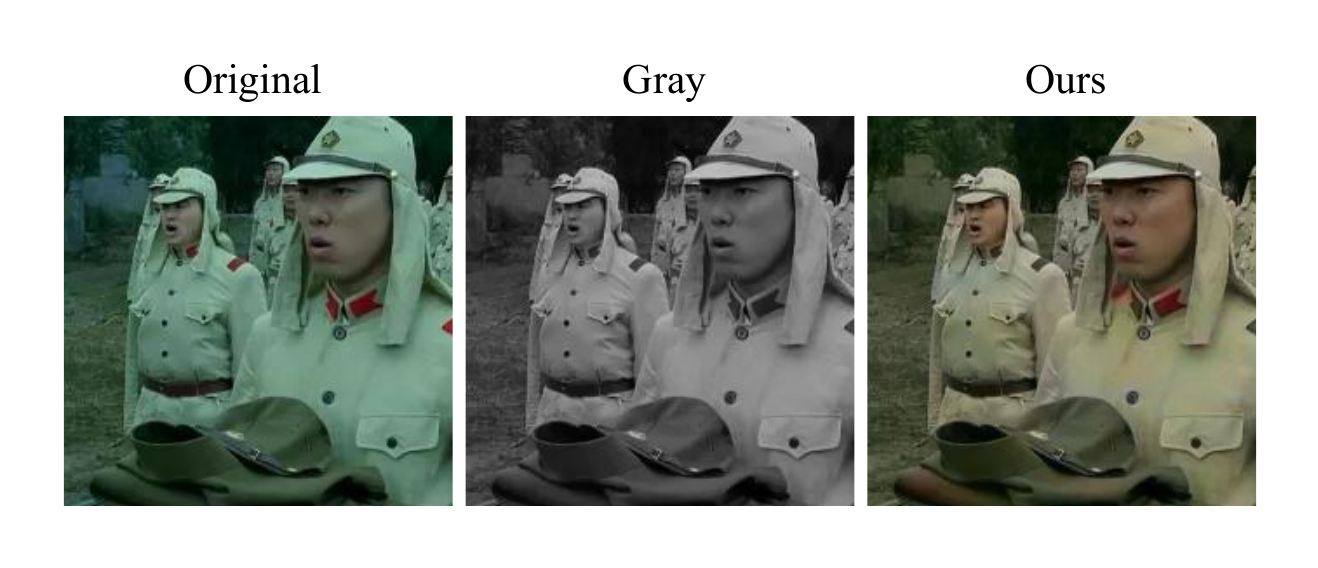}
	\caption{HistoryNet can correct the color and ﬁnally achieve a better colorization result. For example, this historical photo appears yellow and green as a whole.}
	\label{fig6}
\end{figure}

\begin{figure}[htbp]
	\centering
	\includegraphics[width=\linewidth]{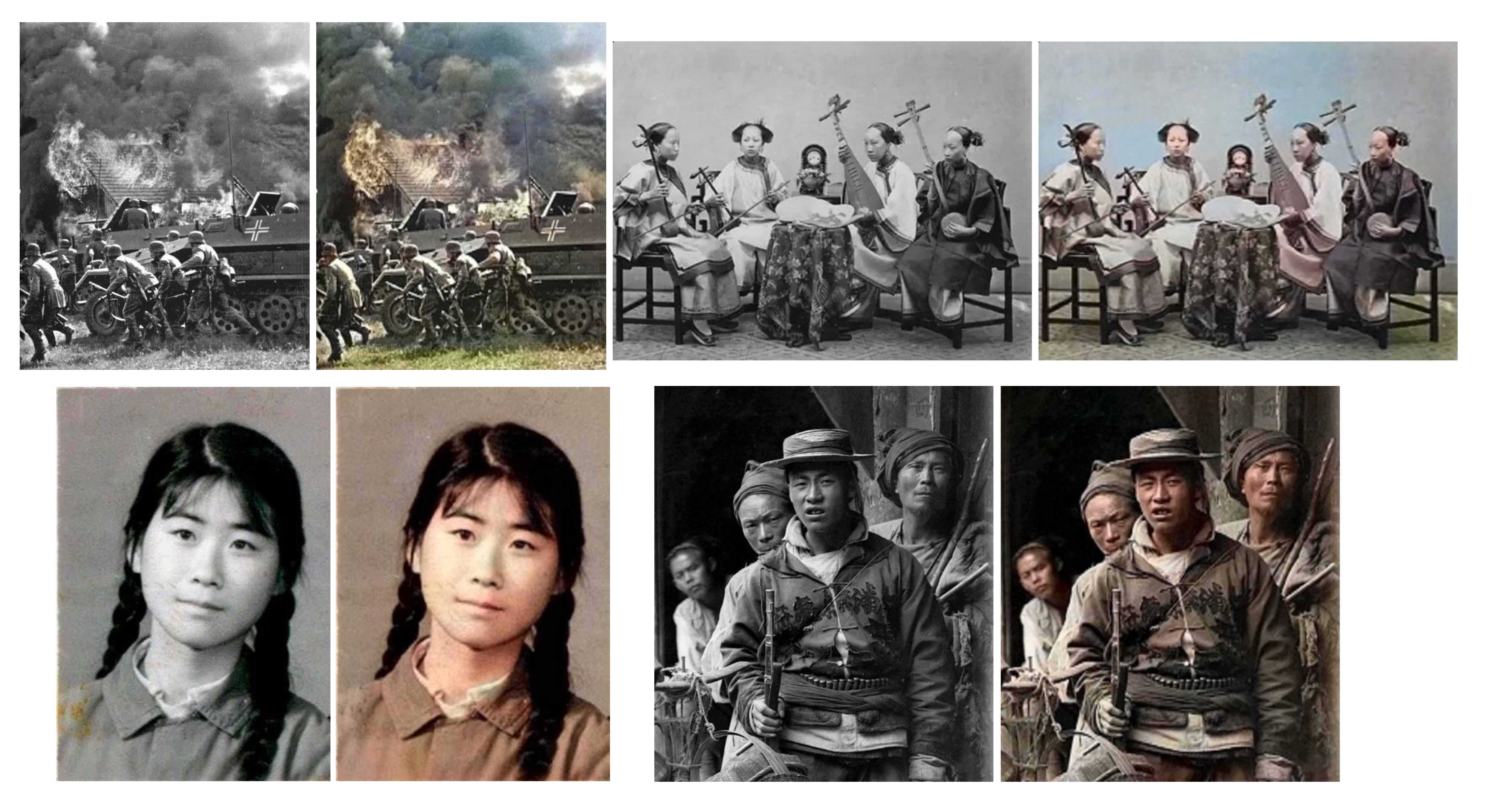}
	\caption{HistoryNet also apply for some legacy black and white images.}
	\label{blackwhite}
\end{figure}

\textbf{Results on old photos} We design MHMD suitable for colorizing historical images, especially military uniforms and suits of historical images. In addition, our method can restore old photos, remove the background noise and colorize the yellowing background to the normal color. As shown in Figure \ref{fig6}. We also colorize some legacy black and white images. As shown in the Figure \ref{blackwhite}, we can see our method is still applicable and has good effect for black and white photos.

\textbf{Comparisons with state-of-the-art} Figure \ref{duibishiyan} shows the experimental results of the five best current methods. It can be seen from the Figure \ref{duibishiyan} that our method has a good performance on the accuracy of the image color and boundary of each part, while other methods have problems such as inaccurate colorizing of the images and discontinuous lumpiness of the color on the clothes. From the first and fourth line of Figure \ref{duibishiyan}, we can see that each element (such as clothes, face etc.) in our colorization results have a clear boundary and achieve the consistency within the region block, that is, the clothes part is of the same color, From this comparative experiment, we can see the advantages of designing parsing and classifier submodules in HistoryNet. Parsing subnetwork can solve the problem of boundary segmentation and classifier subnetwork can achieve the consistency and continuity of the overall coloring of the persons clothing. 

In addition, our method is not only suitable for the colorization of military uniform, but also for the colorization of other natural landscapes and characters, as shown in the Figure \ref{othercomparison}.

\section{Conclusions}
In this paper, we propose a new HistoryNet architecture, which contains parsing, classification and classifier subnetworks. Semantic parsing subnetwork can help the colorization boundary more accurate. Classifier subnetwork can help choose correct color. In addition, we propose a dataset called MHMD that focus on the real gray historical images. To the best of our knowledge, the proposed MHWD dataset is the largest dataset of historical image colorization. The MHWD can be accessed by request. Through relevant qualitative and quantitative comparison, our method is superior to the state-of-the-art colorization network in LPIPS, PSNR, SSIM. Another purpose of our work is to inspire more researchers to make the gray image colorization technologies more useful in historical image/video colorization. In the future, related research methods will also be used in video coloring.

\section*{Acknowledgements}
We thank the ACs, reviewers, and the annotators of our dataset such as Jisen Huang and Xuntao Zhou. This work is partially supported by the National Natural Science Foundation of China (62072014), the Beijing Natural Science Foundation (L192040), and the Advanced Discipline Construction Project of Beijing Universities (20210041Z0401).


\bibliographystyle{ACM-Reference-Format}
\bibliography{sample-base}


\end{document}